\documentclass[letterpaper,twocolumn,10pt]{article}
\usepackage{usenix2019_v3}
\usepackage{tikz}
\usepackage{amsmath}
\usepackage{multirow}
\usepackage{makecell}

\usepackage{filecontents}
\usepackage{subcaption}
\usepackage{xspace}
\usepackage{booktabs}
\usepackage{xurl}
\usepackage{wrapfig}
\usepackage[framemethod=TikZ]{mdframed}
\usepackage{tcolorbox}
\newtcolorbox{mybox}{colback=red!5!white,colframe=red!75!black}
\usepackage{flushend}

\mdfdefinestyle{MyFrame}{
    linecolor=black,
    outerlinewidth=0.5pt,
    roundcorner=5pt,
    innertopmargin=2pt,
    innerbottommargin=2pt,
    innerrightmargin=-10pt,
    innerleftmargin=-15pt,
    leftmargin = -10pt,
    rightmargin = -10pt,
    backgroundcolor=gray!10,
    userdefinedwidth=220pt
}

\usepackage{subcaption}
\begin{document}

\date{}

\definecolor{ForestGreen}{RGB}{34,139,34}

\newcommand{\approach}{Contrastive Textual Deviation\xspace}
\newcommand{\abbrv}{CTD\xspace}

\title{Enabling Contextual Soft Moderation on Social Media\\ through \approach\thanks{This paper is accepted for publication at the 2024 USENIX Security Symposium. Please cite accordingly.}}

\author{Pujan Paudel, Mohammad Hammas Saeed, Rebecca Auger, Chris Wells, and Gianluca Stringhini\\
Boston University\\
\{ppaudel,hammas,raauger,cfwells,gian\}@bu.edu
}

\maketitle
\newcommand{\descr}[1]{\smallskip\noindent\textbf{#1}}
\newcommand{\descrit}[1]{\smallskip\noindent\emph{#1}}

\begin{abstract}

Automated soft moderation systems are unable to ascertain if a post supports or refutes a false claim, resulting in a large number of contextual false positives.
This limits their effectiveness, for example undermining trust in health experts by adding warnings to their posts or resorting to vague warnings instead of granular fact-checks, which result in desensitizing users.
In this paper, we propose to incorporate stance detection into existing automated soft-moderation pipelines, with the goal of ruling out contextual false positives and providing more precise recommendations for social media content that should receive warnings.
We develop a textual deviation task called \approach (\abbrv) and show that it outperforms existing stance detection approaches when applied to soft moderation.
We then integrate \abbrv into the state-of-the-art system for automated soft moderation Lambretta, showing that our approach can reduce contextual false positives from 20\% to 2.1\%, providing another important building block towards deploying reliable automated soft moderation tools on social media.

\end{abstract}

\maketitle

\section{Introduction}

\begin{figure*}[t]
     \centering
     \begin{subfigure}[b]{0.33\textwidth}
         \centering
         \includegraphics[width=\textwidth]{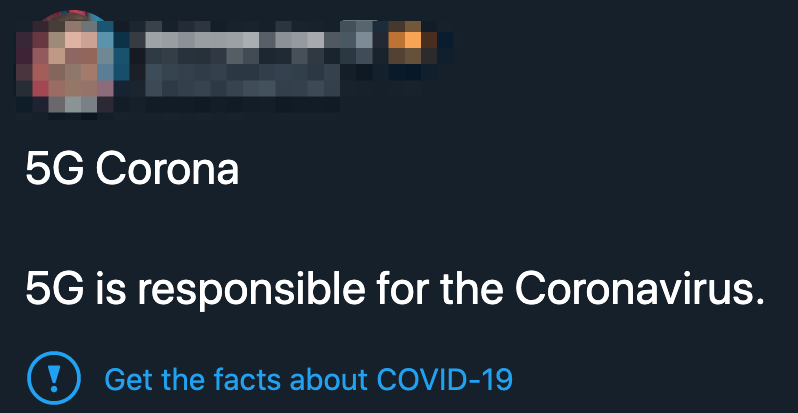}
         \caption{Correct application of warning label.}
         \label{fig:example_correct}
     \end{subfigure}
     \begin{subfigure}[b]{0.33\textwidth}
         \centering
         \includegraphics[width=\textwidth]{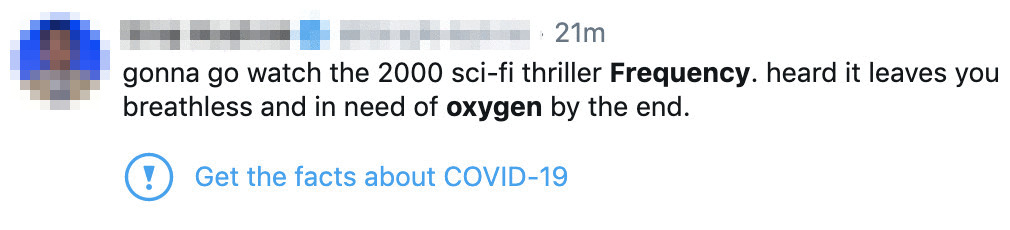}
         \caption{Topical False Positive.}
         \label{fig:example_topical_falsepositive}
     \end{subfigure}
     \begin{subfigure}[b]{0.33\textwidth}
         \centering
         \includegraphics[width=\textwidth]{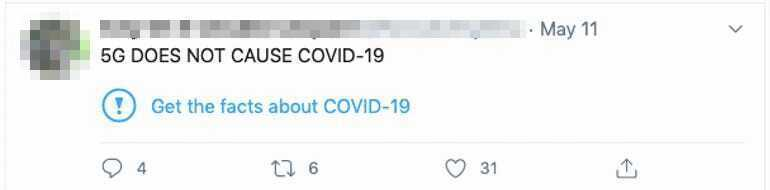}
         \caption{Contextual False Positive.}
         \label{fig:example_contextual_falsepositive}
     \end{subfigure}
        \caption{Three tweets discussing the debunked claim that COVID-19 is caused by 5G. Existing moderation systems might suffer from topical false positives as well as contextual false positives.}
        \label{fig:stance_motivator}
\end{figure*}

``\emph{Get the facts about COVID-19}.''
This message became a familiar sight for the users of Twitter during the pandemic surge in 2020.
Added as a warning label by the platform to accompany potentially misleading tweets about COVID-19 (see Figure~\ref{fig:example_correct}), its goal was to allow users to seek more information and not fall for dangerous conspiracy theories and misinformation which could have compromised public safety~\cite{sharevski2021misinformation}.
This was an early example of \emph{soft moderation}, which was later applied by Twitter on other important topics (e.g., the 2020 US Presidential election~\cite{zannettou2021won}) and has since been adopted by other platforms like Facebook~\cite{softmoderation_facebook}, Instagram~\cite{softmoderation_instagram}, and TikTok~\cite{ling2022learn}.

Platforms do not release details on how they identify tweets that need to receive soft moderation labels, but this includes getting input from reputable authorities like fact-checking organizations (e.g., Snopes) and public agencies (e.g., the CDC), applying automated tools like semantic matching and information retrieval to identify posts about known misleading claims, and relying on human moderators to vet content that should receive a warning~\cite{humanmoderators_pipeline}.
Despite these efforts, however, recent work has showed that the methods adopted by online platforms fail to flag a large amount of misleading content that should be moderated~\cite{ling2022learn,paudel2022lambretta,zannettou2021won}.

Fully automated attempts by platforms to identify misleading content had shortcomings.
We illustrate an example of this in Figure~\ref{fig:stance_motivator}.
To curb the spread of a conspiracy theory linking 5G technology to the spread of COVID-19, Twitter decided to add a warning label to any tweet that contained the words ``\emph{5G}'' and ``\emph{oxygen}''~\cite{businessinsider2020}.
Due to the generality of these keywords, this caused the platform to flag tweets that were unrelated to the conspiracy theory, like the one in Figure~\ref{fig:example_topical_falsepositive}.
We call these irrelevant tweets mistakenly flagged for moderation \emph{topical false positives}.
In addition to irrelevant tweets, Twitter also applied warning labels to tweets that debunked the conspiracy theory, like the one in Figure~\ref{fig:example_contextual_falsepositive}~\cite{cnet2020}.
This can have adverse effects, for example by having the platform undermine the public trust in experts (e.g., health professionals) by mistakenly flagging their informative posts as misleading~\cite{wapo2022}.
It would also discourage diligent users who actively debunk misinformation on social media (i.e., ``wisdom of the crowds''~\cite{pennycook2019fighting,miyazaki2023fake}), who would see their fact-checks marked as misleading.
Finally, it can cause \emph{warning fatigue}~\cite{morrow2022emerging} in social media users, where users are bombarded with warning labels attached to the posts and stop paying attention to them, thereby reducing the intended effectiveness of the labels.
We call this type of mistakenly flagged posts \emph{contextual false positives}.
We argue that an effective approach for the automated identification of content that should receive moderation should address both topical and contextual false positives.

Recently, the computer security community has focused on developing automated systems to flag content that should receive moderation labels.
Lambretta~\cite{paudel2022lambretta} is a system that leverages information retrieval techniques (i.e., Learning to Rank) to identify the optimal set of keywords that represent a misleading claim and identify social media posts that are discussing that claim.
Lambretta outperforms alternative keyword retrieval and semantic search approaches, and largely addresses the problems of topical false positives, reporting a false detection rate of 3.93\% on a dataset of tweets discussing the 2020 US Presidential Election.
However, Lambretta does not address the problem of contextual false positives, being unable to ascertain if a social media post is supporting a false claim (and therefore should be moderated) or refuting it.
In fact, the false detection rate from the perspective of contextual false positives on its original dataset is 20\%.
This limitation is not unique to Lambretta, but is also present in other content moderation systems powered by retrieval or semantic similarity approaches~\cite{shao2016hoaxy,gaglani2020unsupervised}.

To reduce contextual false positives, \emph{stance detection} techniques can be integrated into soft moderation systems.
Stance detection is a task where the objective is to learn a model that can identify whether a given piece of text is in favor of or against a set of target(s)~\cite{aldayel2021stance}.
In the context of fact-checking, the problem aims to automatically detect if a piece of text supports or refutes a misleading claim~\cite{hardalov2022survey}.
We find, however, that existing stance detection techniques are limited in their ability to generalize across unseen claims~\cite{ng2022my,schiller2021stance} and different social media platforms~\cite{mccoy2019right}, making them unfit to be applied on social media platforms at scale.

To address the shortcomings of existing stance detection approaches when dealing with content moderation on social media, we reframe the problem as a textual deviation detection one and propose \emph{\approach} (\abbrv).
Our approach consists of developing, for each claim that we wish to moderate, a triplet composed of a \textcolor{blue}{consensus statement} that is confirmed to be true (e.g., by health authorities or fact-checking organizations) and two contrastive markers, a piece of \textcolor{red}{refuting evidence} and one of \textcolor{ForestGreen}{supporting evidence}.
Our approach then leverages the \emph{emergent} abilities of Large Language Models (LLMs) like zero-shot learning to identify if a piece of text supports or refutes the consensus statement.

Compared to conventional stance detection approaches, our method has two major advantages.
First, we build a model that can identify patterns of deviation from an anchor text (i.e., the \textcolor{blue}{consensus statement}) in a topic-invariant fashion, without depending on features specific to a certain topic (e.g., climate change denial or vaccine misinformation).
This overcomes one of the main issues of traditional supervised stance detection pipelines, where researchers need to annotate samples for each claim as these methods fail to generalize well across multiple claims~\cite{ng2022my,schiller2021stance}.
Second, \abbrv can leverage the consensus statements curated by fact checkers and expert authorities as part of the stance detection process (through the use of the \textcolor{blue}{consensus statement}), allowing us to keep experts in the loop in the case of evolving events like a global pandemic or misinformation stories that threaten public safety. 

We first motivate the task of textual deviation detection by identifying the shortcomings of existing approaches for stance detection in the context of content moderation.
We then perform preliminary experiments on the advantages of reframing stance detection as textual deviation by bootstrapping the task using the zero-shot learning capabilities of Large Language Models (LLMs).
Motivated by the better performance of LLMs over alternative solutions and by the ease of generalizing across datasets, we further fine-tune the LLM for the task of \abbrv to boost its performance.
Finally, we integrate our fine-tuned unsupervised stance detection model into the analysis pipeline of the state-of-the-art soft moderation system Lambretta~\cite{paudel2022lambretta}, to improve its detection capabilities.
We make all the labeled datasets curated in the work and fine-tuned models publicly available \footnote{\url{https://huggingface.co/collections/ppaudel/contrastive-textual-deviation-65e20c48680724cc9a809062}}.

In summary, this paper makes the following contributions:
\begin{itemize}
    \item We propose \approach (\abbrv) as a new task to perform stance detection on social media. 
        We show that \abbrv overcomes the limitations of existing stance detection approaches and that it generalizes across platforms by testing it on datasets collected from different platforms (Twitter and Reddit) and covering different topics (COVID-19, climate change, and politics).
    \item We fine-tune a LLM dedicated to the task of \abbrv and show that our approach improves over baseline methods, with an average improvement on the F1-score of over 7\%. 
    \item We integrate \abbrv into the automated soft moderation pipeline of the state-of-the-art approach Lambretta, reducing contextual false positives from 20\% to 2.1\%, with a minimum impact on false negatives.
\end{itemize}

\section{Datasets}
\label{sec:dataset}
We curate and collect multiple datasets throughout our work that i)~come from multiple social media platforms (Reddit, Twitter), ii)~cover different topics (e.g, Climate Change, COVID-19), and iii)~contain different levels of granularity (i.e, fine-grained or coarse-grained claims).
A summary of the datasets used in our work is presented in Table~\ref{tab:datasetsummary}.

\begin{table*}[t]
\centering
\begin{tabular}{|l|l|l|l|l|l|}
\hline
\textbf{Dataset} & \textbf{Platform} & \textbf{Topic} & \textbf{\#Claims} & \textbf{\#Refute} & \textbf{\#Support} \\ \hline
GWSD~\cite{luo2020detecting} & News & Climate & N/A & 400 & 777 \\ \hline
Climate Skepticism & Reddit & Climate & 3 & 1,277 & 1,650 \\ \hline
COVID-FACT~\cite{saakyan2021covid} & Reddit & COVID-19 & 4,086 & 2,790 & 1,296 \\ \hline
COVID-CQ~\cite{mutlu2020stance} & Twitter & COVID-19 & 1 & 3,488 & 3,515 \\ \hline
Stanceosaurus~\cite{zheng2022stanceosaurus} & Twitter & Fact-Checking & 190 & 1,442 & 3,025 \\ \hline
Election Denial~\cite{paudel2022lambretta} & Twitter & Election & 3 & 128 & 454 \\ \hline
PERSPECTRUM~\cite{chen2019seeing} & Debates & Argument Mining & 907 & 2,468 & 2,627 \\ \hline
\end{tabular}
\caption{Summary of datasets used.}
\label{tab:datasetsummary}
\end{table*}

\descr{Global Warming Stance Dataset (GWSD)~\cite{luo2020detecting}.}
This dataset contains 1,177 ``opinion spans'' of news headlines about Global Warming annotated by human annotators as either accepting of global warming (e.g., 
``\emph{We can't afford to wait until everyone is feeling the pain of the climate emergency before we do something about it}'') or skeptical of it (e.g., ``\emph{Humans have negligible impact on the climate}'').
The opinion spans are extracted from 56,000 news articles spanning a period of 20 years from news outlets such as the New York Times, Fox, Washington Post, Forbes, etc.
We characterize this dataset as a ``coarse-grained'' stance dataset since the opinion spans in this dataset discuss climate change from the broader perspective of climate emergency and skepticism, and do not deal with granular causes or effects of climate change (e.g., \emph{Urban Heat Island effect on Global Warming is negligible}.)
This dataset is slightly imbalanced in its class labels as it contains 777 opinion spans supporting climate change 
and 400 opinion spans skeptical of climate change.

\descr{Climate Skepticism.}
The majority of stance detection datasets are topic-based or target-based in nature, with a very limited set of datasets on claim-based stance detection occurring on social media text. 
This existing gap motivates us to prepare a comprehensive claim-based stance detection dataset.
We curate a dataset of climate change denial discussions, containing Reddit posts that support or refute three different claims related to different arguments used in climate change discussions.
We query the Pushshift Reddit dataset~\cite{baumgartner2020pushshift} with curated keywords related to the claims, retrieving posts from Reddit that are discussing narratives related to three of the most popular climate denier claims: i)~\emph{Cosmic rays are causing global warming (GW)}, ii)~\emph{Antarctica is gaining ice}, and iii)~\emph{Urban Heat island (UHI) effect exaggerate global warming trends}.
These claims are fine-grained and objective in nature, unlike high-level ``topics'' in the GWSD dataset such as ``Global warming is a hoax,'' or ``Climate change is not happening.''~\cite{mohammad2016semeval,luo2020detecting}
For the goal of scalable and tractable soft moderation, these claims are very representative of misleading claims propagating on social media that can be easily refuted from scientific consensus, or authoritative sources (e.g. Skeptical Science). 
We then set out to annotate 1,000 Reddit posts for each claim by developing a codebook dedicated to climate skepticism. 
Our codebook was informed by the crowdsourced resource Skeptical Science, which provides pointers for understanding different narratives used by climate skeptics to deny a claim as well as examples of scientific support for confirmed claims.
We do not include posts that are inquiring about the claim in question, are neutral towards it, or are topically irrelevant.
Two researchers performed multiple rounds of annotation and reached a near-perfect agreement of $\kappa$ = 0.865~\cite{mchugh2012interrater}) Cohen's Kappa (i.e., strong agreement).
This dataset also has value in evaluating the transferability of stance detection approaches as it contains multiple targets (claims), and different types of label distribution within the claims.
A summary of the climate denial claims and the number of annotated Reddit posts for each claim is provided in Table~\ref{tab:dataset_climate_eval}.

\begin{table}[]
\centering
\begin{tabular}{|l|l|l|}

\hline
\textbf{Claim}                & \textbf{\#Refute} & \textbf{\#Support} \\ \hline
Cosmic Rays cause GW& 577                & 439                \\ \hline
UHI effect exaggerate GW trends     & 390                & 495                \\ \hline
Antarctica is gaining ice         & 683                & 343                \\ \hline
\end{tabular}
\caption{Summary of climate skeptic claims annotated with stance labels.}
    \label{tab:dataset_climate_eval} 
\end{table}

\descr{COVID-FACT~\cite{saakyan2021covid}.}
This dataset belongs to a family of datasets in the field of automated fact-checking known as FEVER (Fact Extraction and Verification).
What makes it interesting for our work is that the FEVER dataset contains labeled text known as ``evidences'' that either supports or refutes a given claim.
More precisely, the COVID-FACT dataset contains 4,086 claims concerning the COVID-19 pandemic, and both refuting (3,488) and supporting evidence (3,515) for the claims.
These claims are obtained from the \textit{/r/COVID19} community of Reddit and are ``fine-grained'' in nature.

\descr{COVID-CQ~\cite{mutlu2020stance}.}
In addition to creating our own dataset for identifying stance in climate denial claims, we also make use of an existing stance detection dataset of COVID-19 treatment tweets. 
The dataset, called COVID-CQ, contains 7,003 tweets and their respective stance on the efficacy of hydroxychloroquine as a treatment for COVID-19.
The annotation criteria of this work closely align with the annotation criteria of our own data collection process, and this dataset fits the problem definition of claim-based stance detection very closely.

\descr{Stanceosaurus~\cite{zheng2022stanceosaurus}.}
This dataset is a multi-lingual and multi-cultural corpus of tweets annotated with a stance towards 4,467 tweets discussing 190 misinformation claims from 9 fact-checking sources (Snopes, Poynter, FullFact, etc.)
We use a subset of the dataset containing fact-checked claims and tweets in English, which either support or refute the claims, ignoring the tweets that are either \textit{irrelevant} to the claim, or simply \textit{querying} about the claim.
For the posts that are \textit{discussing} the claims in question, the dataset also contains annotation if the tweets are leaning towards supporting or refuting the claim, and we use tweets that are annotated as such with the label of their leaning.

\descr{Election denial~\cite{paudel2022lambretta}.}
This dataset contains 499 election denial claims and 101,353 tweets discussing the claims retrieved by the corresponding soft-moderation system Lambretta. 
We select the three most popular election denial claims by frequency present in the dataset and annotate a sample of 200 random tweets discussing the three claims for the stance of the tweets concerning the claim in question.
We adapt the codebook and annotation process used for annotating climate denial-related claims.
A near-perfect agreement of $\kappa$ = 0.866~\cite{mchugh2012interrater} was reached among the annotators.
A summary of the claims, and the number of supporting, and refuting tweets for each claim is presented in Table~\ref{tab:dataset_lambretta_eval}.

\begin{table}[]
    \scalebox{0.8}{
\begin{tabular}{|l|l|l|}
\hline
\textbf{Claim}                & \textbf{\#Refute} & \textbf{\#Support} \\ \hline
Wisconsin Voter Turnout above 90\% & 132                & 32                \\ \hline
Illegal suitcase of ballots in Georgia       & 161& 55                \\ \hline
Dead Voters voted in Michigan           & 161                & 41                \\ \hline
\end{tabular}
    }
\caption{Summary of election denial claims from ~\cite{paudel2022lambretta} annotated with stance labels}
    \label{tab:dataset_lambretta_eval}
\end{table}

\descr{PERSPECTRUM~\cite{chen2019seeing}.}
Finally, we use a dataset from argument mining called PERSPECTRUM~\cite{chen2019seeing}.
PERSPECTRUM contains 907 claims from online debate topics and 5,095 ``perspectives'' from search engine results presenting diversifying viewpoints about the claims.
These ``perspectives'' are annotated with stance as either supporting or refuting the claim and cover more than 10 different topics such as Politics, Freedom of Speech, Environment, Science, Health etc.
The structure of this dataset provides value in augmenting a large-scale stance dataset as it contains claims and multiple sides (perspectives) of the claims from different topics.

\descr{Summary.} A summary of all the datasets curated and collected is listed in Table~\ref{tab:datasetsummary}.
We can observe that the datasets used by our work span across different use cases of misinformation such as climate denial, public health emergencies, civic processes such as elections, and general purpose fact-checking.
This way, we aim to comprehensively evaluate our method on a variety of misleading claims occurring across two different social media platforms.

\section{Motivation: Existing stance detection methods fall short}
\label{sec:motivation}

We identify three requirements that effective stance detection approaches should satisfy to successfully improve content moderation systems.
Ideally, we would be able to leverage existing stance detection methods for this purpose.
However, we find that previous work, including more sophisticated entailment-based methods~\cite{hanselowski2018ukp}, fall short in achieving one or more of these requirements.
In a nutshell, the requirements and shortcomings that we identify are the following:

\begin{mybox}
\begin{itemize}
    \item \textbf{R1. Need for Granularity}: Supervised methods trained on coarse-grained claims fail on fine-grained claims.
    \item \textbf{R2. Need for Claim Invariancy}: Supervised methods trained on one claim do not generalize well on other claims.
    \item \textbf{R3. Need for Contrastive Context Awareness}: Supervised entailment-based methods fail to identify stance on social media despite being given context-aware hypothesis statements.
\end{itemize}
\end{mybox}

In the rest of this section, we discuss the shortcomings of existing approaches in achieving the three requirements, showcasing our preliminary experiments.
Then, based on these observations, in Section~\ref{sec:bootstrapping} we present our solution, which overcomes the limitations of previous methods. 

\subsection{Need for Granularity (R1)}
\label{sec:motivation_specificity}

While coarse-grained claims (e.g., ``\emph{Climate change isn't real}'') may generally describe broad categories of misinformation, finer-grained ones (e.g., ``\emph{Cosmic rays are causing global warming}'') are often encountered online.
An effective stance detection approach should be able to operate on both categories of claims.

To investigate the ability of stance detection to generalize between coarse-grained and fine-grained claims, we use the GWSD and Climate Skepticism datasets.
These two datasets both pertain to climate change, but GWSD contains coarse-grained claims while the Climate Skepticism one is built from fine-grained ones.
We train a supervised model on the GWSD dataset and test whether this model can identify the stance of climate skepticism on the fine-grained claims of the Climate Skepticism dataset.
Following the standard BERT fine-tuning recipe~\cite{rogers2021primer}, we fine-tune a DistilBERT~\cite{sanh2019distilbert} model for five epochs with a learning rate of $5e^{-5}$ until the training loss converges.
We balance the dataset by randomly undersampling the supporting claims, as the original dataset is imbalanced. 
We repeat the experiment five times and report the average evaluation of the fine-tuned BERT model on the three claims from the Climate Skepticism dataset.
\begin{table}[]
\center
\begin{tabular}{|l|l|}
\hline
\textbf{Claim} & \textbf{F1} \\ \hline
Antarctica is gaining ice & 0.61 \\ \hline
UHI exaggerate GW trends & 0.44 \\ \hline
Cosmic rays cause GW & 0.41 \\ \hline
\end{tabular}
\caption{Performance of BERT model on GWSD claims evaluated on Climate Skepticism dataset.}
\label{tab:result_gswd_bert}
\end{table}

\descr{Takeaways.} The fine-tuning procedure on the GWSD dataset performs well when cross-evaluated on coarser claims (F1 score of 0.766 on 5-fold cross-validation).
However, the model performance drops drastically when evaluated on the fine-grained climate claims from the Climate Skepticism dataset as seen in Table~\ref{tab:result_gswd_bert}.
This experiment shows that supervised models trained on coarse-level claims fail to evaluate the stance on fine-grained ones, despite having high domain and topic overlap with the training data.
This motivates the need to design stance detection methods that are highly granular and specific in nature.

\subsection{Need for Claim Invariancy (R2)}
\label{sec:motivation_invariancy}
The second requirement that we identify is that a stance detection approach trained on a fine-grained claim about a topic must generalize to detect stance on other fine-grained claims about the same topic.
This is important because false information is not static and new claims emerge all the time (e.g., the emergence of narratives advocating for Vitamin C, Hydroxychloroquine, and Ivermectin as effective cures against COVID-19 at different points in time during the pandemic).

To investigate if existing stance detection approaches can generalize between claims, we again use the Climate Skepticism dataset, as it contains three fine-grained claims.
It is to note that all three corpora for the individual claims come from the same social media platform (Reddit), thus we can expect the corpus distribution of the evaluation setting to be similar to the training setting.
As in the previous step, we train a DistilBERT model for each claim and evaluate the model on detecting the stance of posts on the other two claims that it was not trained on.
The results are provided in Table~\ref{tab:result_claim_invariancy}.

\descr{Takeaways.} We find that in all three cases, a model trained on one claim fails to generalize on other claims as the performance drops drastically.
This motivates the need for building claim invariant stance detection methods that are not only learning features to detect stance specific to the training data of the claim it is trained on, but generic representations of support or refute towards a claim.

\begin{table}[]
\center
\scalebox{0.9}{
\begin{tabular}{|l|l|l|l|}
\hline
\textbf{Training Claim} & \textbf{F1 \#1} & \textbf{F1 \#2} & \textbf{F1 \#3} \\ \hline
\#1 Antarctica is gaining ice     & \textbf{0.828}        & 0.429                 & 0.551                 \\ \hline
\#2 UHI exaggerate GW trends   & 0.470                 & \textbf{0.7315}       & 0.578                 \\ \hline
\#3 Cosmic rays cause GW         & 0.4183                & 0.589                 & \textbf{0.817}        \\ \hline
\end{tabular}
}
\caption{Stance detection performance across claims. Each line shows the F1 score for the model trained on one claim and tested on the three claims in the Climate Skepticism dataset.}
\label{tab:result_claim_invariancy}
\end{table}

\subsection{Need for contrastive context awareness (R3)}
\label{sec:motivation_entailment}

An effective stance detection approach must be able to accurately identify text that supports or refutes a given claim.
A promising approach to achieve this is Natural Language Inference (NLI), which is also known as Recognizing Textual Entailment (RTE).
NLI is a widely popular approach for detecting stance in NLP, where a claim is treated as a \emph{premise} and a piece of evidence is treated as a \emph{hypothesis}. 
The task then consists in checking if the premise entails (supports) or contradicts (refutes) the hypothesis~\cite{hanselowski2018ukp}.
Compared to the approach of fine-tuning BERT-based models for classification tasks, NLI can be adapted with a much more granular objective for stance detection, as each claim being evaluated can be directly subjected to the most relevant piece of evidence associated with the claim.
One of the major reasons existing stance detection methods fail to generalize well on detecting stance on a new topic during inference is the lack of enough contextual information about the topic or target they are subjected in the out-of-domain or zero-shot setting~\cite{jiang2023zero,burnham2023stance}.
Prior research showed that providing enough context about the topic or claim being evaluated can bridge this context gap, improving stance detection models to generalize well on unseen topics~\cite{beck2023robust}.
Based on this, we investigate if we can leverage NLI methods for stance detection on social media by providing the most ``context-aware'' set of evidence (hypothesis) statements related to a claim (premise) getting evaluated.
Our assumption is that if there is \emph{sufficient} context provided in the \emph{hypothesis} statement, NLI models should be able to detect stance well on new claims and topics.

We first train an NLI model on COVID-19 claims using the COVID-FACT dataset.
We then evaluate the trained model on the COVID-CQ dataset.
The FEVER setup of the COVID-FACT dataset is perfect for this task, as it already contains pieces of evidence supporting or refuting a claim.
In addition to that, these claims are ``context-aware,'' (e.g., about the lack of studies confirming the effectiveness of Hydroxychloroquine to treat COVID-19) instead of being generic like ``Covid-19 and mask'' or ``Covid-19 and HCQ.''
Also, COVID-FACT consists of claims and evidence that are semantically close to those being evaluated in the COVID-CQ dataset (e.g., ``homemade remedies of COVID-19'' or ``alternative treatments of COVID-19''), therefore we can expect a high level of domain overlap, similar granularity, and semantic overlap between the training data and the evaluation data.

Following a similar methodology as in prior sections, we train a DistilBERT model for textual entailment using these pairs and evaluate the trained model on the COVID-CQ dataset.
Since NLI only allows to specify one hypothesis for each premise, we evaluate the trained model on two different configurations using the following hypotheses: i)~Another study has confirmed hydroxychloroquine to be effective in the treatment of COVID-19 (coronavirus), ii)~No clinical studies have confirmed hydroxychloroquine as a cure for COVID-19 (coronavirus).
We then consider a tweet as supporting the misleading narrative if it either entails with first hypothesis or contradicts the second hypothesis.
Similarly, a tweet debunks the misleading narrative if it either contradicts the first hypothesis or entails the second hypothesis.
This way, we setup an evaluation setting for a model that is trained on context-aware sentence pairs, and evaluate it under an identical setting, allowing us to understand if satisfying the criteria of ``context-awareness'' is sufficient enough for stance detection for content moderation.

\descr{Takeaways.} We evaluate the trained DistilBERT model on the COVID-CQ dataset and find that the best-performing F1 score of this method is 0.53, which is only slightly better than random chance, despite having a high topical overlap with the training dataset, and further taking a granular approach of training entailment or contradiction aligned on claims.
Moreover, we discover that the NLI model evaluated on the ``context-aware'' setting, providing the best set of hypothesis statements with ``sufficient'' information about the claim, is not adequate for stance detection on social media text.
These results suggest that providing a context-aware hypothesis statement is not enough to build NLI models for precise stance detection. 
We argue that while context awareness through the best set of hypothesis statements gives a model important contextual signals about the claim, it would be beneficial for the model to be ``contrastively context-aware,'' i.e., exposed to contradicting hypotheses, one supporting and one refuting the claim, which is what commonly occurs on social media.

\section{\approach}
\label{sec:bootstrapping}

Based on the three requirements defined in the previous section, we aim to build an unsupervised stance detection model that overcomes the limitations of previous approaches by: i)~detecting stance at a fine-grained level, ii)~learning semantic representations of stance on a claim invariant fashion, and iii)~encoding contrastive context awareness to learn higher level representations of stance.
We first describe our proposed solution by formally defining a new task called \approach (\abbrv) and discuss the different components of the task.
We then bootstrap \abbrv by leveraging the zero-shot learning capabilities of Large Language Models (LLMs) through prompt engineering.
Finally, we conclude the section by demonstrating the successful performance of bootstrapped \abbrv by comparing against multiple baselines from supervised learning. 
\subsection{Task Definition}

We propose to reframe the problem of stance detection as a new task of ``textual deviation'' detection, which aims to satisfy all three requirements.
First, for every misleading claim that we wish to moderate, we start from a \textcolor{blue}{\emph{consensus statement}} that has been confirmed to be true, for example by health authorities like the World Health Organization (WHO) in cases of public health emergencies or by fact-checking organizations in case of political events.
Our intuition is that posts that refute the claim, spreading false information while doing so, will deviate from the consensus statement, while those that support (debunk) the claim will stick closer to the messaging of the consensus statement.
We then extend the idea of ``context aware'' textual entailment to ``contrastively context-aware'' textual entailment by providing a pair of ``contrastive markers'' for each consensus statement associated with the misleading claim: a piece of \textcolor{red}{refuting evidence} and one of \textcolor{ForestGreen}{supporting evidence}.
Our idea is to leverage the zero-shot capabilities of Large Language Models (LLMs) to align social media posts discussing a claim with the piece of anchor ``consensus'' statement specific to the claim, while using the ``contrastive markers'' as additional context to aide the alignment decision.
We call this task \emph{\approach} (\abbrv).

\begin{figure}[t]
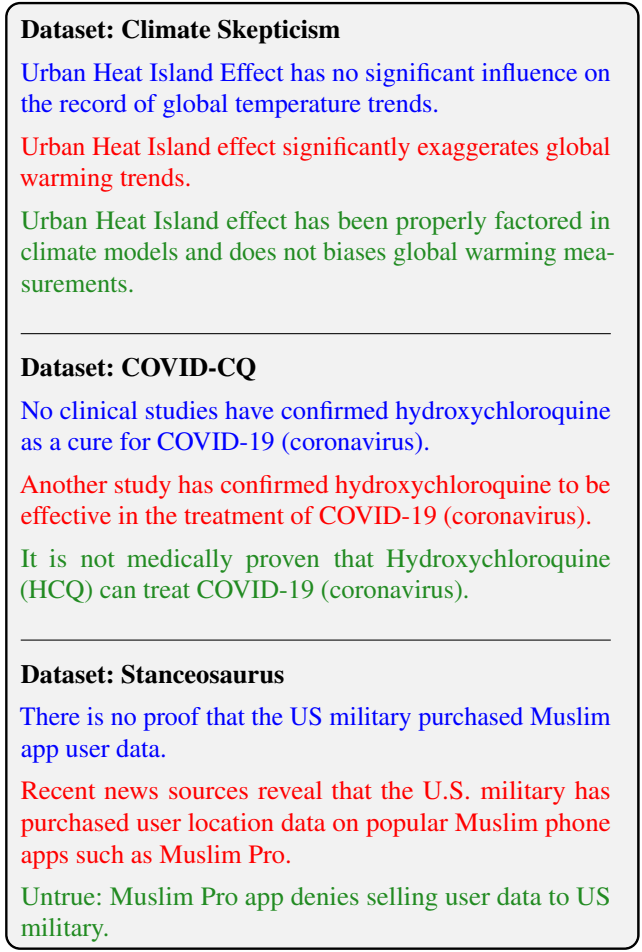

\begin{mdframed}[style=MyFrame,nobreak=true]
\begin{quote}

\textbf{Dataset: Climate Skepticism}

\textcolor{blue}{Urban Heat Island Effect has no significant influence on the record of global temperature trends.}

\textcolor{red}{Urban Heat Island effect significantly exaggerates global warming trends.}

\textcolor{ForestGreen}{Urban Heat Island effect has been properly factored in climate models and does not biases global warming measurements.}

\rule{\linewidth}{0.4pt}

\textbf{Dataset: COVID-CQ}

\textcolor{blue}{No clinical studies have confirmed hydroxychloroquine as a cure for COVID-19 (coronavirus).}

\textcolor{red}{Another study has confirmed hydroxychloroquine to be effective in the treatment of COVID-19 (coronavirus).}

\textcolor{ForestGreen}{It is not medically proven that Hydroxychloroquine (HCQ) can treat COVID-19 (coronavirus).}

\rule{\linewidth}{0.4pt}

\textbf{Dataset: Stanceosaurus}

\textcolor{blue}{There is no proof that the US military purchased Muslim app user data.}

\textcolor{red}{Recent news sources reveal that the U.S. military has purchased user location data on popular Muslim phone apps such as Muslim Pro.}

\textcolor{ForestGreen}{Untrue: Muslim Pro app denies selling user data to US military.}
\end{quote}
\end{mdframed}
\caption{Example claims from evaluation dataset and corresponding triplets.}
\label{fig:example}
\end{figure}

The core of \abbrv lies on formulating this triplet of i)~a \textcolor{blue}{consensus statement}, ii)~a \textcolor{red}{refuting evidence}, and iii)~ a \textcolor{ForestGreen}{supporting evidence} for each claim that we want to infer the stance of social media posts.
Compared to traditional supervised learning approaches which need hundreds or even thousands of examples for training claim-specific or topic-specific stance models, we argue that curating this triplet is a much easier and \emph{tractable} human effort, which could be easily carried out by fact-checking organizations or social media platforms.
The \textcolor{blue}{consensus statements} used to anchor the triplets are the ``truth values'' about the false claim in question.
For each false claim in the datasets, we use the corresponding fact-checks and scientific explanations associated with the claim.
For Climate Skepticism claims, we use fact-checks from \url{SkepticalScience.com} available under the ``\textit{What the science says}'' snippet for every climate skepticism claim.
Stanceosaurus contains fact-checked claims coming from several reputable sources (e.g. PolitiFact, Snopes, etc.), and we used the debunking article headline from the corresponding fact-checking sources as the consensus statement.
For the COVID-CQ dataset, we used Google Fact Check Explorer to identify fact-checks and used their debunking title as the consensus statement.
To complete the triplet for each claim, we formulate the contrastive markers as a positive affirmation and a negative reframing of the corresponding \textcolor{blue}{consensus statement}.
Figure~\ref{fig:example} shows examples of triplets from three different claims from our evaluation datasets, with \textcolor{blue}{consensus statement}, \textcolor{ForestGreen}{supporting evidence}, and \textcolor{red}{refuting evidence} for each of the claims.

\begin{figure}
    \includegraphics[width=\columnwidth]{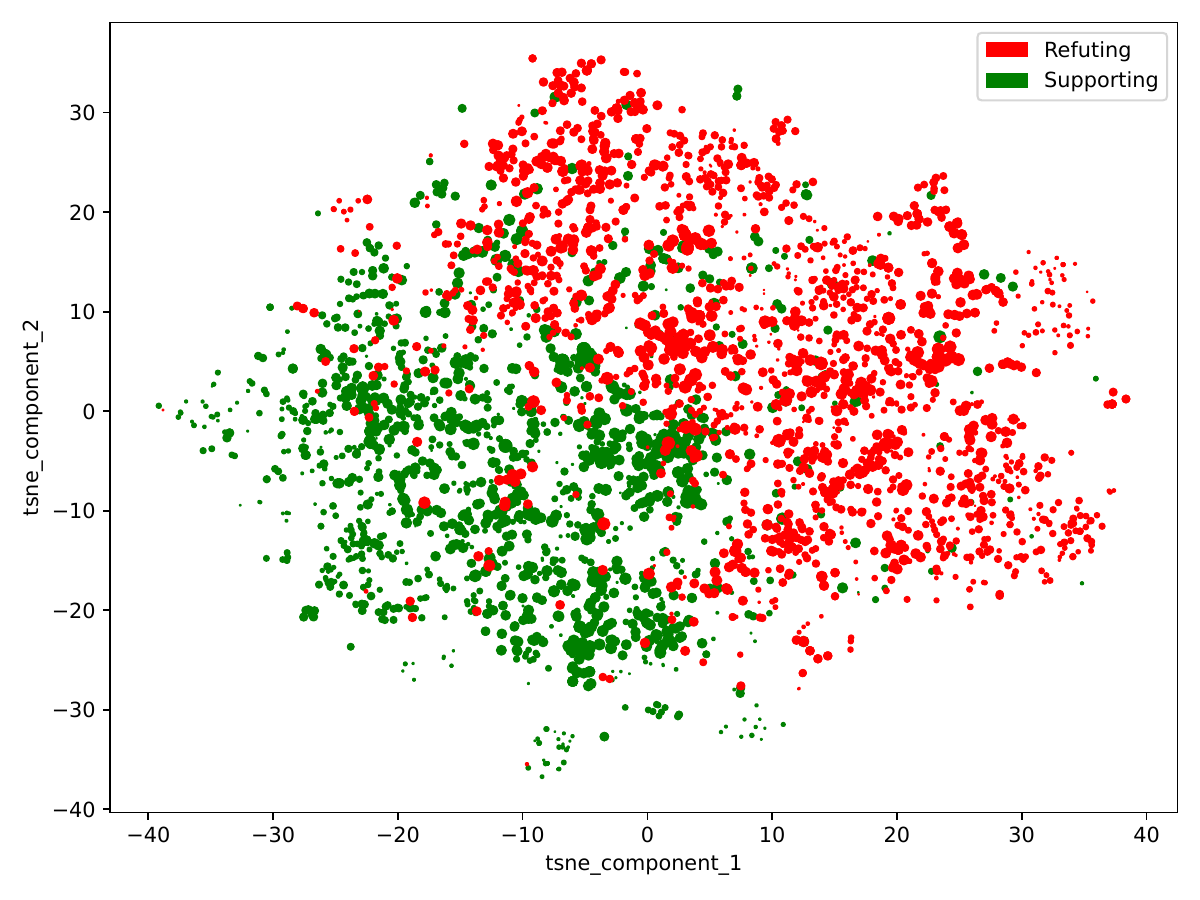}
\caption{T-SNE embeddings of supporting and refuting statements from COVID-CQ}
\label{fig:tsne}
\end{figure}

\abbrv allows us to satisfy all three requirements discussed in Section~\ref{sec:motivation}.
R1 is addressed by allowing us to use customizable and well-informed \textcolor{blue}{consensus statements} directly discussing the claim being moderated.
To address R2, \abbrv learns semantic features of deviation from an anchor piece of text (\textcolor{blue}{consensus statements}), rather than semantic features of support or refute towards a particular claim that the model is trained on, allowing us to scale our method across claims and achieve the property of claim invariancy.
Finally, \abbrv addresses R3 by using a pair of contrasting evidence as context input for the classification task (\textcolor{red}{refuting evidence} and \textcolor{ForestGreen}{supporting evidence}).
This allows our approach to improve over entailment-based methods, which only use one piece of hypothesis statement and thus do not generalize to the case of content moderation on social media, as we showed in Section~\ref{sec:motivation_entailment}.
Before bootstrapping \approach as a NLP task, we empirically validate the underlying intuition motivating \abbrv by visualizing if tweets supporting a claim are semantically different than tweets refuting it.
We utilize the tweets from the COVID-CQ dataset for this experiment and embed the tweets using the Sentence-T5 encoder~\cite{ni2022sentence}, producing 768-dimensional embeddings.
Note that these sentence embeddings are produced using the same encoder component of the LLM we will use for bootstrapping purposes i.e. (FLAN-T5-XXL), and thus serve as a useful tool for empirically testing our intuition.
To this end, we center the embeddings of tweets from the COVID-CQ dataset using the embedding of the \textcolor{blue}{consensus statement} and project the T-SNE embeddings of the supporting and refuting tweets.
The scatterplot in Figure~\ref{fig:tsne} shows a strong separation between the supporting and refuting tweets, with minimal overlap, indicating that the supporting tweets and refuting tweets are indeed semantically different. 
Additionally, we also compute the cosine similarity of the embeddings of the supporting tweets and refuting tweets with the \textcolor{blue}{consensus} statement.
Statistical analysis using a two-sample t-test to compare the means of the supporting and refuting tweets allows us to reject the null hypothesis (with $p <<< 0.01 $ and $t-statistics >>> 0$).
This further validates our intuition that the posts supporting the claim align more with the consensus than the posts that refute the claim.

\subsection{Bootstrapping \abbrv using LLMs}
Due to their zero-shot learning abilities~\cite{wei2021finetuned}, Large Language Models (LLMs) can be used off-the-shelf to leverage the triplet structure defined by \abbrv to detect stance in an unsupervised fashion.
We argue that LLMs are particularly well suited for this task because: i)~they have demonstrated performance competitive to supervised baselines in multiple NLP tasks~\cite{wei2021finetuned,puri2019zero}, ii)~they can incorporate the proposed contrastive markers through in-context learning~\cite{wei2022emergent}.

\descr{Background.} Recent approaches showed that \emph{prompting} a LLM to a specific \emph{prompt} that defines the downstream task (e.g., sentiment analysis, textual entailment), and the possible label spaces for the tasks (e.g., positive, negative, or entailment, contradiction) can lead to performance comparable to supervised methods for text classification~\cite{puri2019zero}.
This ability of LLMs has been shown to be ``emergent:'' as the size and scale of these language models increases, so does their ability to generalize to unseen tasks~\cite{wei2022emergent}.
Another emergent ability of LLMs is their capability to perform ``in-context learning.''
By extending the \emph{prompts} to include a few examples of the downstream task and the classification labels, LLMs can learn new tasks.
Subjecting LLMs to demonstrative examples can give the model an idea of the label space, of the mapping between input and label space, and help generalize LLMs to a variety of tasks without the need for task-specific fine-tuning. 
In-context learning does not require any model update or fine-tuning and performs competitively with state-of-the-art methods for text classification problems~\cite{brown2020language}.

\descr{Prompt engineering to bootstrap LLMs.} 
To show the feasibility of our approach for soft moderation, we first bootstrap \abbrv with LLMs.
To this end, we build our prompts as follows:
For every social media post discussing a claim, we subject it to a \textcolor{blue}{consensus statement} that explains the established consensus (e.g., a fact-check or a statement on the scientific consensus about the claim).
Then, we proceed to show an example statement that \textcolor{red}{refutes} the consensus. 
Similarly, we show an example statement that \textcolor{ForestGreen}{supports} the consensus.
Finally, we end the prompt with the text that we wish to classify using the LLM, where the LLM is expected to return one of the stance labels (support, or refute).
Figure~\ref{fig:prompt_setup} illustrates the prompt structure that we use for bootstrapping \abbrv.

\begin{figure}
\begin{mdframed}[style=MyFrame,nobreak=true]
\begin{quote}
Classify if a statement supports or refutes the consensus statement: \textcolor{blue}{<Consensus statement>.}

Statement: \textcolor{red}{<Refuting evidence>.}

Response: Refutes.

Statement:  \textcolor{ForestGreen}{<Supporting evidence>.}

Response: Supports.

Statement: <Text from the test sample to classify.>

Response:
\end{quote}
\end{mdframed}
\caption{Prompting structure for bootstrapping \abbrv.}
\label{fig:prompt_setup}
\end{figure}

Our intuition is that by doing so we can leverage the zero-shot classification capabilities of LLMs, while at the same time \emph{ground} them towards incorporating the consensus and the contrasting evidence as additional context for classification.

\descr{Validating \abbrv bootstrapping.}
We use the open-source encoder-decoder model FLAN-T5 by Google for bootstrapping purposes~\cite{chung2022scaling}, 
prompting, for each claim, the LLM with the prompt structure defined above.
The underlying architecture of FLAN-T5 is a T5 model (Text-to-Text Transfer Transformer), a new unified paradigm of transfer learning for text classification.
By formulating challenging tasks like machine translation, summarization, etc. to be a text-to-text task, T5 has shown promising results in multiple NLP benchmark problems over encoder-based models like BERT~\cite{roberts2019exploring}.
FLAN-T5 is an improved version of the T5 model that has been ``instruction finetuned'' on more than 1,000 NLP tasks and thus can be used to perform zero-shot text classification by just ``prompting'' the model.

As a summary of our validation results, we present the results of the task bootstrapping on the Climate Skepticism dataset.
We will comprehensively evaluate our task with a fine-tuned model against all three evaluation datasets and claims later in Section~\ref{sec:evaluation}.

As baselines, we compare our bootstrapped task against two different prompt setups: i)~LLM without contrastive markers, ii)~LLM without consensus definition.
Additionally, we also compare our method against two supervised methods for stance detection i)~a DistilBERT~\cite{sanh2019distilbert} model fine-tuned on each claim, and ii)~a fine-tuned NLI model on ClimateFever dataset~\cite{diggelmann2020climate} (following the identical procedure we fine-tuned NLI model on COVID-FACT dataset in Section~\ref{sec:motivation_entailment}).
The purpose of evaluating the two added prompt configurations alongside the prompt configuration for \abbrv is to understand the role that each component of the triplets (consensus statement and the contrastive markers) plays in identifying stance.
The prompt configuration of evaluating LLMs without contrastive markers will help us understand if the contrastive markers are an improvement over entailment-based approaches.
For the DistilBERT model, which uses supervised fine-tuning, we perform 5-fold cross-validation for each of the claims and report the average score across the folds.
We use the weighted F1 score as the evaluation metric as the label distribution of supporting and refuting labels for each of the three claims in the Climate Skepticism dataset is imbalanced.

\begin{table}[h]
    \centering

  \begin{tabular}{|l|l|}
\hline
\textbf{Method}             & \textbf{Mean F1} \\ \hline
BERT Fine Tuning            & 0.786       \\ \hline
Climate Fever NLI           & 0.626       \\ \hline
LLM w/o Contrastive Markers & 0.795           \\ \hline
LLM w/o Consensus           & 0.771          \\ \hline
\textbf{Bootstrapped CTD}            & \textbf{0.836}       \\ \hline
\end{tabular}

    \caption{Summary of F1 Scores for Different Methods}
    \label{tab:bootstrapping_results}
\end{table}

\descr{Results.} A summary of the bootstrapping results with the mean F1 scores for the three Climate Skepticism claims is presented in Table~\ref{tab:bootstrapping_results}.
We find that our bootstrapped method (which is unsupervised) performs better than supervised BERT-based fine-tuning baselines.
Additionally, we find that our approach provides better performance over using the LLMs in their default state: without consensus grounding (LLM without consensus), and without contrastive markers (LLM without contrastive markers).
The results also suggest that the triplet setup of \abbrv performs better than the conventional setup of entailment for stance detection.
Additionally, we observe that having the contrastive markers by themselves without the consensus statement (LLM without consensus) is not nearly as effective for detecting stance.
Overall, these results validate that the task of \abbrv can outperform existing methods for supervised stance detection.
Moreover, the utility of our approach is validated in being fully unsupervised and requiring minimal setup needed for a claim (i.e. a \textcolor{blue}{consensus} statement, and a pair of \textcolor{ForestGreen}{supporting} and \textcolor{red}{refuting} in-context examples).

\section{Evaluation}
\label{sec:evaluation}

Following the successful bootstrapping of \abbrv through LLMs, we now further fine-tune the FLAN-T5 model to improve upon the bootstrapping performance and build a robust model for the task.
First, we discuss the setup of this fine-tuning procedure and the process of augmenting an argument-mining dataset for the task of \abbrv.
Next, we comprehensively evaluate the performance of this fine-tuned model against the prior bootstrapped LLM and various other baselines for the downstream task of stance detection in all three of our evaluation datasets.
We then assess how the runtime and detection performance of CTD varies with increasing model size.
Finally, we integrate our approach with the state-of-the-art soft moderation pipeline Lambretta~\cite{paudel2022lambretta}, showing that it can improve its ability to limit contextual false positives.

\subsection{Fine tuning FLAN-T5 for \abbrv}
\label{sec:finetuning}

The bootstrapping experiments from Section~\ref{sec:bootstrapping} suggest that \abbrv can be a reliable approach to identify if a social media post is supporting or refuting a claim, enabling better content moderation.
While LLMs bootstrapped on this task already perform better than multiple existing baselines by leveraging in-context learning, we can improve them further by fine-tuning LLMs that are specifically dedicated to the task of \abbrv.
Previous work showed that even with little training data fine-tuned LLMs perform better on tasks~\cite{roberts2019exploring,liu2022few,tinn2023fine}.
Additionally, fine-tuning can potentially allow the LLM to build a more specialized model representation associated with the \abbrv task, enabling the model to better generalize and also be adversarially robust~\cite{nookala2023adversarial}.
Previous work showed that the performance of LLMs can be heavily affected by the components in the prompt, in our case the triplets used by \abbrv.
For this reason, we aim to use the large number of triplet combinations available in our fine-tuning dataset, to learn a model that is decoupled from the phrasing of the prompt, and more focused and consistent on the \abbrv task itself.

There are two different approaches to fine-tuning LLMs: i)~task-adaptive fine-tuning, and ii)~behavioral fine-tuning.
Fine-tuning a LLM directly on the downstream task is called task-adaptive fine-tuning~\cite{ruder2021lmfine-tuning}.
Behavioral fine-tuning aims to teach a model to learn higher-level language representation by fine-tuning on a task that is behaviorally similar, or close to the downstream task, rather than fine-tuning on the downstream task itself.
This approach is also called intermediate-task fine-tuning, and it works best with tasks that require high-level inference and reasoning capabilities~\cite{pruksachatkun2020intermediate,phang2020english}, like \abbrv. 
Thus, we argue that instead of fine-tuning a LLM for the task of stance detection (task-adaptive fine-tuning), which would suffer from similar limitations as BERT fined-tuned on the stance detection task discussed in Section~\ref{sec:motivation}, a LLM fine-tuned for textual deviation on a large number of claims will be able to reliably perform stance detection on general claims. 
Additionally, behavioral fine-tuning on textual deviation tasks solves the problem of the amount of data available for fine-tuning, as the augmentation of argument mining datasets for textual deviation allows us to generate a dataset that is orders of magnitude larger than any dataset available for specific stance detection topics.
This way, our fine-tuned model is learning what \emph{deviation} or \emph{affirmation} to any piece of anchor claim looks like without relying on target-specific or claim-specific features, further helping us to achieve the need for claim invariancy (R2) (see Section~\ref{sec:motivation_invariancy}).

We use the PERSPECTRUM dataset to fine-tune LLMs for the \abbrv task.
This dataset contains both claims and multiple examples of evidences called \emph{perspectives} either agreeing with the claims or disagreeing with them, as required by \abbrv.
We augment this dataset to create a large-scale one (expanding up to millions of elements) containing the triplets for the textual deviation task.
An example of a claim sampled from the PERSPECTRUM dataset alongside two randomly sampled supporting and refuting perspectives to the claim is presented in Figure~\ref{fig:example_perspectrum}.
This particular claim contains 13 supporting perspectives and 22 refuting perspectives in total.
With the claim being used as the anchor (\textcolor{blue}{consensus}) statement, and the contrasting set of \textcolor{red}{refuting} and \textcolor{ForestGreen}{supporting} perspectives, we can generate a combination of 9,438 triplets for this claim alone.
This way, the PERSPECTRUM dataset containing 907 claims yields a total of 3,311,548 triplets that we use for fine-tuning a FLAN-T5 model for the \abbrv task. 

It is to note that the claims present in this dataset are argumentative sentences used for debates and unrelated to the task of fact-checking or misinformation.
We argue that the normalized argumentative structures used to fine-tune the model for \abbrv will help the model to better capture the intricacies of textual deviation from a canonical perspective, allowing \abbrv to generalize well and achieve claim invariancy (R2) when applied in the wild. 
As we will show later in Section~\ref{sec:eval_comprehensive}, the representations learned through these triplets built from normalized argumentative structures indeed generalize very well in out-of-domain data on Reddit posts and tweets, confirming that higher-level language representations are learned via behavioral fine-tuning.

\begin{figure}[t]
\begin{mdframed}[style=MyFrame,nobreak=true]
\begin{quote}

\textbf{Claim: \textcolor{blue}{All countries should have the right to pursue a nuclear defence.}}

\rule{\linewidth}{0.1pt}

Supporting Perspective \#1: \textcolor{ForestGreen}{All countries are entitled to self defense with nuclear weapons, even when they do not have the capacity to carry conventional weapons.}

Supporting Perspective \#2: \textcolor{ForestGreen}{The pursuit of nuclear defence (respectively the possession of nuclear weapons) by more countries is a guarantee for peace.}

\rule{\linewidth}{0.4pt}

Refuting Perspective \#1: \textcolor{red}{The threat of a state developing nuclear weapons could instigate pre-emptive strikes from its neighbours and rivals to prevent the acquisition of such weapons.}

Refuting Perspective \#2: \textcolor{red}{It is very difficult to intercede in humanitarian crises in states wherein nuclear weapons are present.}

\end{quote}
\end{mdframed}
\caption{Example claims and perspectives from PERSPECTRUM dataset.}
\label{fig:example_perspectrum}
\end{figure}

We fine-tune the FLAN-T5 model using a Parameter Efficient Fine-tuning (PEFT) technique known as LoRA (Low-Rank domain adaptation)~\cite{hu2021lora}.
LLMs like FLAN-T5 contain billions of parameters (11B) and fine-tuning the entire network is not feasible even under expensive GPU memory configurations.
LoRA works by freezing the model weights of LLMs and injecting trainable rank decomposition matrices into each layer of the transformers, greatly reducing the number of trainable parameters for downstream tasks while achieving comparable or even better performance~\cite{hu2023llm}.
In our case, LoRA reduces the number of trainable parameters of the 11B FLAN-T5 LLM by 84\% (from 11B to 18.8M).

Then, using the PEFT configuration of LoRA, we fine-tune the FLAN-T5 model on the augmented fine-tuning dataset.
To make the task more manageable, we randomly sample 4 contrastive examples per claim to fine-tune the model, resulting in a dataset size of 36,000 triplets from 905 claims.
We divide the fine-tuning dataset into 85-15 training and validation splits and fine-tune the FLAN-T5 model for 5 epochs until the validation loss stops decreasing further.
Finally, we test this fine-tuned FLAN model on all three of our evaluation datasets (one coming from Reddit and two from Twitter).
The results of this experiment are shown in Table~\ref{tab:finetuning_results}.
As it can be seen, the fine-tuned \abbrv model performs well, reporting F1-scores between 0.84 and 0.90 on the different datasets.

\subsection{Comparison with existing baselines}
\label{sec:eval_comprehensive}

We further evaluate the fine-tuned FLAN-T5 model against three types of baselines: task fine-tuned BERT, multi-task deep neural networks, and behaviorally fine-tuned baselines.

\descr{Task Fine-tuned BERT.} This is a supervised learning setup where the BERT model is directly trained on a fraction of the evaluation dataset in a five-fold cross-validation setup.
The evaluation configuration for the Climate Skepticism and COVID-CQ datasets has already been discussed in Section~\ref{sec:motivation}.
It is to note that the third evaluation dataset we use, Stanceosaurus contains a large number of claims, and only a few tweets (in the order of 100) discussing a single claim.
Fine-tuning a BERT-based model for classification for each claim with a limited number of tweets per claim is not possible, thus we fine-tune a BERT-based model for textual entailment instead (identical to NLI model trained on COVID-Fact in Section~\ref{sec:motivation_entailment}).
We refer to this setup as ``Task-BERT.''

\descr{Multi-Task Deep Neural Networks (MT-DNN) based stance detection.} 
A promising solution to DNN methods generalizing poorly outside their training domain (as demonstrated in Section~\ref{sec:motivation_invariancy}) is a framework known as Multi-Task Deep Neural Network (MT-DNN)~\cite{liu2019multi}, where a variety of datasets from different domains are used to train a model to increase its robustness.
This has helped in learning powerful higher-level representations across multiple Natural Language Understanding (NLU) tasks, and obtain state-of-the-art results in them.
The MT-DNN framework has been adapted for stance detection by Schiller et al.~\cite{schiller2021stance}, reporting improvement in stance detection performance.
We use the best-performing MT-DNN model available in their work that is fine-tuned across 10 different datasets as a baseline.

\descr{Behaviorally fine-tuned baselines.} The concept of using contrastive markers and an anchor statement proposed in our work is very closely associated with a framework in Deep Learning known as Triplet learning~\cite{schroff2015facenet}, which has been successfully applied in computer vision tasks~\cite{zeng2020hierarchical} and NLP tasks~\cite{reimers2019sentence}.
The fine-tuned BERT model we evaluated and compared on sequence classification (in Section~\ref{sec:motivation_invariancy}) and sentence-pair classification (in Section~\ref{sec:motivation_entailment}) are pre-trained with the objective of Masked Language Modeling (MLM), and do not consider any type of contrastive loss in their design.
However, a recent framework known as POLITICS (Pretraining Objective Leveraging Inter-article Triplet-loss using Ideological Content and Story)~\cite{liu2022politics} leverages the concept of Triplet Loss~\cite{schroff2015facenet} to pre-train a language model.
This model modifies BERT to be trained on a large corpus of news articles discussing the same story, but from different ideologies and uses triplet loss to capture the ideological (dis)similarity among articles on the same story~\cite{liu2022politics}.
Furthermore, POLITICS has been fine-tuned for downstream tasks like stance detection, showing improved performance on various datasets over conventional BERT models.
We therefore use POLITICS as an additional baseline to compare against \abbrv.

POLITICS needs to be fine-tuned on downstream tasks to be useful.
To this end, we fine-tune the POLITICS model on the same dataset we fine-tuned the FLAN-T5 model on (PERSPECTRUM), adopting Natural Language Inference (NLI) as the downstream task setup.
Similar to how we extracted triplets from PERSPECTRUM in Section~\ref{sec:finetuning}, we generated combinations of around 11K pairs of claims and perspectives for fine-tuning the POLITICS model.
The process of fine-tuning the POLITICS model closely mirrors that of the FLAN-T5 model, except for the PEFT configuration, which was not necessary for POLITICS.

\begin{table}[t]
    \centering
    \begin{tabular}{|l|l|l|}
        \hline
        \textbf{Claim / Dataset} & \textbf{Method} & \textbf{F1} \\
        \hline
        \multirow{5}{*}{Climate Skepticism (Reddit)} & Task-BERT & 0.786 \\
        & MT-DNN & 0.645 \\
        & POLITICS & 0.734 \\
        & Bootstrapped \abbrv & 0.836 \\
        & \textbf{Fine-tuned \abbrv} & \textbf{0.871}\\
        \hline
        \multirow{5}{*}{COVID-CQ (Twitter)} &  Task-BERT & 0.900 \\
        & MT-DNN & 0.586 \\
        & POLITICS & 0.831 \\
        & Bootstrapped \abbrv & 0.810\\
        & \textbf{Fine-tuned \abbrv} & \textbf{0.904} \\
        \hline
        \multirow{5}{*}{Stanceosaurus (Twitter)} & Task-BERT & 0.731 \\
        & MT-DNN & 0.656 \\
        & POLITICS & 0.670 \\
        & Bootstrapped \abbrv & 0.773\\
        & \textbf{Fine-tuned \abbrv} & \textbf{0.848} \\
        \hline
    \end{tabular}
    \caption{Results of comprehensive evaluation.}
    \label{tab:finetuning_results}
\end{table}

\begin{table}[]
\begin{tabular}{|l|l|l|l|}
\hline
\textbf{Model} & \textbf{\# params} & \textbf{Runtime (s)} & \textbf{Mean F1} \\ \hline
FLAN-T5-Small  & 60M                 & 0.008                 & 0.492            \\ \hline
FLAN-T5-Base   & 250M                & 0.012                 & 0.555           \\ \hline
FLAN-T5-Large  & 780M                & 0.018                 & 0.589          \\ \hline
FLAN-T5-XL     & 3B                  & 0.043                 & 0.811            \\ \hline
\textbf{FLAN-T5-XXL}    & 11B                 & 0.078                 & 0.874            \\ \hline
\end{tabular}
\caption{Performance of different FLAN-T5 models on \approach.}
\label{tab:scaling_laws}
\end{table}

\descr{Results.} 
The summary of evaluation results of the fine-tuned FLAN model and other methods is presented in Table~\ref{tab:finetuning_results}.
We can observe that the fine-tuned FLAN model for \abbrv performs consistently better than all other baselines on all evaluation datasets, outperforming not only Task-BERT and other baselines but also the results from the bootstrapped \abbrv. 
First, this confirms our hypothesis that the zero-shot learning capabilities of LLMs can be further bolstered by behaviorally fine-tuning a LLM on the \abbrv task.
Secondly, fine-tuned \abbrv has better performance than other stance detection approaches such as MT-DNN and POLITICS while being fine-tuned on the same dataset.
This proves that the task formulation of \approach leveraged by the fine-tuned \abbrv model meaningfully outperforms existing systems for stance detection.
Additionally, it is interesting to note that the FLAN-T5 model, which was fine-tuned on normalized argumentative structures that are domain and platform-independent (as discussed in Section~\ref{sec:finetuning}), consistently outperforms the Task-BERT models that were specifically fine-tuned on claim specific posts from social media text. 
This again affirms our hypothesis that a granular, topic-independent, and platform-independent stance detection model can be created by reframing the problem as a task of \approach, which captures the semantics of stance detection much better on a foundation level.
In summary, these results show that the FLAN-T5 model fine-tuned on the \abbrv task is a reliable, robust, and scalable stance detection method for the soft moderation of social media text.

\subsection{Model size and performance tradeoff}

Finally, we study how the runtime and performance of \abbrv varies based on the different models available in the FLAN-T5 family.
Following the same methodology discussed in Section~\ref{sec:finetuning}, we fine-tune four smaller models of \abbrv : i)~FLAN-T5-Small, ii)~FLAN-T5-Base, iii)~FLAN-T5-Large, and iv)~FLAN-T5-XL. 
This helps us better understand the scaling properties of the \abbrv task.
Moreover, researchers and practitioners can use this to guide them in appropriate model selection based on their resource constraints and runtime requirements.
Table~\ref{tab:scaling_laws} shows that the performance of \abbrv increases linearly as the model size increases along with the tradeoff on runtime.
On the other hand, \abbrv fine-tuned on FLAN-T5-XL can be a viable option for practitioners as the performance dropoff from the best model (i.e. FLAN-T5-XXL) is relatively minor compared to the substantial reduction in model size (11B to 3B).

\subsection{Integrating \abbrv into Lambretta}
\label{sec:ctdwithlambretta}

At this point, we have experimentally validated that a fine-tuned FLAN model on the \abbrv task has better performance than existing supervised baselines and LLMs bootstrapped for the task, making it the state-of-the-art approach in claim-based stance detection.
Our motivation for designing \abbrv, however, is to enable soft moderation approaches to get rid of contextual false positives, allowing platforms to deploy more effective warnings that are only applied to content that is supporting a certain false claim.
The state-of-the-art soft moderation tool Lambretta~\cite{paudel2022lambretta} is unable to discern this contextual information, and while it performs well in discarding candidate posts that are irrelevant to a given claim, it still flags a large fraction of posts that refute false claims as candidates for moderation: looking at the results reported by Lambretta on three false claims related to the 2020 US Presidential Election (see Table~\ref{tab:dataset_lambretta_eval}), we find that 20\% of the candidates flagged by the system are contextual false positives.

\begin{figure}
    \includegraphics[width=\columnwidth]{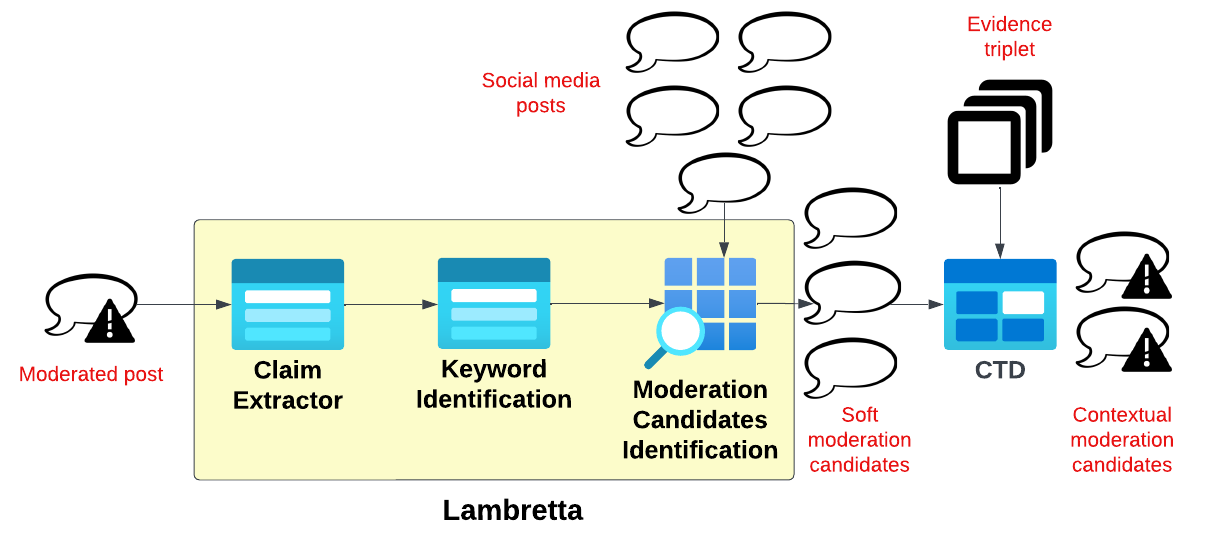}
\caption{Our envisioned integration of \abbrv as a downstream component to Lambretta's soft moderation system.}
\label{fig:lambretta_ctd}
\end{figure}

To evaluate how much our approach can aid existing soft moderation approaches to incorporate contextual knowledge into their decisions, we add \abbrv as a downstream component to Lambretta~\cite{paudel2022lambretta}, and use its stance detection capabilities to further refine the posts that it flags as candidates for soft moderation, filtering out posts that have a debunking (supporting) stance.
Our goal is to reduce contextual false positives while keeping false negatives at a minimum.
Our envisioned architecture is showed in Figure~\ref{fig:lambretta_ctd}.
It is to note that while we evaluate the approach on Lambretta, \abbrv can be added as a post-retrieval filter in any content moderation system after posts relevant to a claim or event for flagging are retrieved.

\descr{Setup.} 
We use the annotated dataset of election denial tweets discussed in Table~\ref{tab:dataset_lambretta_eval} for this purpose.
These tweets were flagged by Lambretta as part of the evaluation in the original paper~\cite{paudel2022lambretta}.
We can observe that the class of stance in the tweets is heavily skewed towards spreading the misinformation (\textbf{\textit{Refuting}} the fact-checks), and contains a smaller proportion of tweets that are debunking the misinformation (\textbf{\textit{Supporting}} the fact-checks).
As discussed, Lambretta has no contextual understanding of its candidates for soft moderation, and when evaluating for contextual false positives, we find that the system reports a 20\% False Detection Rate on average, with an F1 score on the three claims that ranges between 0.88 and 0.89 depending on the claim (see Table~\ref{tab:eval_lambretta}).
Note that the False Negative Rate for the Lambretta baseline is 0 since the system does not perform any contextual filtering on the moderation candidates, and all retrieved tweets are considered matches for moderation.

\begin{figure}[t]
\begin{mdframed}[style=MyFrame,nobreak=true]
\begin{quote}

\textbf{Claim: Wisconsin voter turnout above 90\%}

\textcolor{blue}{The voter turnout in Wisconsin is within historical averages of 73\% and does not indicate any voter fraud.}

\textcolor{red}{Fraud in Wisconsin as there were more votes than registered voters and irregular voter turnout around 90\%.}

\textcolor{ForestGreen}{Wisconsin did not have more votes than people registered and their voter turnout figures is 73\%.}

\rule{\linewidth}{0.4pt}

\textbf{Claim: Illegal suitcase of ballots in Georgia.}

\textcolor{blue}{Recently circulating viral video on social media doesn’t show ‘suitcases’ of illegal ballots in Georgia.}

\textcolor{red}{Suitcases filled with illegal ballots were pulled out from underneath tables after election observers left in Georgia.}

\textcolor{ForestGreen}{Officials have confirmed that there were no suitcases full of illegal ballots counted in the absence of election observers.}

\rule{\linewidth}{0.4pt}

\textbf{Claim: Dead voters voted in Michigan.}

\textcolor{blue}{There is no credible evidence that dead people voted in the election or ballots were cast fraudulently by deceased voters.}

\textcolor{red}{There were many cases of voter fraud nationwide due to dead people's votes getting counted.}

\textcolor{ForestGreen}{No evidence that 14,000 dead people cast ballots in Wayne County, Michigan.}

\end{quote}
\end{mdframed}
\caption{Triplets used in our experiments on integrating \abbrv with Lambretta.}
\label{fig:triplets_eval}
\end{figure}

To evaluate \abbrv on the candidate tweets flagged by Lambretta, we follow the same methodology discussed in Section~\ref{sec:bootstrapping}, curating a triplet of \textcolor{blue}{consensus statement}, \textcolor{red}{supporting evidence}, and \textcolor{ForestGreen}{refuting evidence} for each of the three election denial claims being evaluated, as showed in Figure~\ref{fig:triplets_eval}.
Note that in deployment, the only input that platform moderators need to use for our system is the triplet structure pertaining to a specific social media claim that they wish to moderate.
In the case of a single social media post being the starting point, moderators need to first extract the claim contained in the post, which can be done by leveraging Lambretta's Claim Extraction component or other claim extraction tools such as OpenIE~\cite{etzioni2008open}.
We then use the fine-tuned FLAN-T5 model presented in Section~\ref{sec:finetuning} with the three curated triplets to identify the stance on the tweets discussing the three claims.
Note that for this experiment \abbrv is not fine-tuned on the tweets flagged by Lambretta, nor on any tweet discussing election fraud claims, further showcasing that our approach is context agnostic and generalizes across different topics.

\descr{Results.} 
The summary of the evaluation is presented in Table~\ref{tab:eval_lambretta}.
As it can be seen, integrating \abbrv with Lambretta largely reduces the rate of contextual false positives, bringing down the average false detection rate by an order of magnitude from 20\% to 2.1\%.
For the ``Wisconsin voter turnout above 90\%,'' the false detection rate after applying our approach is actually zero.
At the same time, the false negative rate remains small, being 1.8\% on average.
This translates in F1 scores between 0.98 and 0.96, showing an improvement of about 10\% over the baseline, showing that \abbrv can be effectively used to improve soft moderation systems for social media.

\begin{table}[t]
\scalebox{0.75}{
\begin{tabular}{|l|l|l|l|l|}
\hline
\textbf{Claim} & \textbf{Method} & \textbf{F1} & \textbf{FDR} & \textbf{FNR} \\ \hline
\multirow{2}{*}{GA suitcase of ballots} & Lambretta & 0.877 & 0.219 & \textbf{0} \\ \cline{2-5} 
 & \textbf{Lambretta + \abbrv} & \textbf{0.987} & \textbf{0.015} & 0.010 \\ \hline
\multirow{2}{*}{Dead Voters voted in MI} & Lambretta & 0.887 & 0.203 & \textbf{0} \\ \cline{2-5} 
 & \textbf{Lambretta + \abbrv} & \textbf{0.9632} & \textbf{0.048} & 0.024 \\ \hline
\multirow{2}{*}{WI Voter Turnout above 90\%} & Lambretta & 0.891 & 0.195 & \textbf{0} \\ \cline{2-5} 
 & \textbf{Lambretta + \abbrv} & \textbf{0.988} & \textbf{0} & 0.022 \\ \hline
\end{tabular}
}
\caption{Evaluation of the end-to-end component.}
    \label{tab:eval_lambretta}
\end{table}

\section{Related Work}
We review related work on stance detection, LLMs for text classification, and use of stance detection for integrated fact-checking.

\descr{Stance detection.}
Stance detection is a foundational technique for various natural language understanding tasks~\cite{wang2018glue} and has been used under various settings like argument mining~\cite{lippi2016argumentation}, rumor detection~\cite{zubiaga2016analysing}, and fake news detection~\cite{pomerleau2017fake}.
The majority of the existing work on stance detection focuses on topic or target-specific stance detection, where they aim to detect the stance of a text towards topic such as ``gun rights,'' ``atheism''~\cite{allaway2021adversarial} and public figures like ``Donald Trump'' or ``Hillary Clinton.''~\cite{mohammad2016semeval}
Only a few works focus on claim-based stance detection~\cite{hossain2020covidlies,thorne2018fever,zubiaga2016analysing}, which is the focus of this work.
Works on claim-based stance detection are mostly geared towards checking facts on formal text like Wikipedia~\cite{diggelmann2020climate,thorne2018fever} or scientific knowledge bases~\cite{wadden2020fact}.
Multiple works have used textual entailment for verifying claims on Wikipedia~\cite{thorne2018fever}, scientific knowledge~\cite{beltagy2019scibert}, and climate change conversations~\cite{diggelmann2020climate}.
There are a few works on claim-based stance detection in multi-lingual settings like Arabic~\cite{baly2018integrating,alhindi2021arastance}, and Crotian~\cite{bovsnjak2019data}.

\descr{LLMs for text classification.}
With the explosion of Large Language Models and rapid development of powerful LLM models like Open AI's GPT-4~\cite{OpenAI2023GPT4TR}, Google's BARD~\cite{google_bard}, Anthropic's Claude~\cite{anthropic_claude} etc., there has been a paradigm shift in approaching text classification problems.
The massive amount of internet-scale data that LLMs see during their pretraining has been harnessed to fine-tune LLM models, producing state-of-the-art results in challenging benchmarks in Natural Language Understanding (NLU)~\cite{roberts2019exploring,wang2018glue,wei2021finetuned}.
In addition to the zero-shot capabilities of LLMs, researchers have also used strategies like in-context learning to improve their performance~\cite{min2022rethinking}, making weakly supervised learning scalable and effective.
This has made bootstrapping LLMs for custom text classification much easier, a process that we demonstrated successfully in our work.
This way, LLMs have presented themselves as a viable alternative for multiple Natural Language Processing tasks such as classification, and summarization, while relaxing the constraint of task-specific training data needed for conventional NLP methods.

\descr{Stance detection as a fact-checking component.}
Prior works have used stance detection as a component of automated fact-checking pipelines.
Works in this domain use stance prediction as one of the many components of their fact-checking pipeline, alongside other components and metadata such as user features and features of conversation threads. 
Zubiaga et al.~\cite{zubiaga2018detection} incorporated stance classification to detect the stance of tweets in a four-step rumor verification pipeline.
Dungs et al.~\cite{dungs2018can} leveraged the stance of conversational threads to predict veracity of rumors.
CredEye, a system proposed by Popat et al.~\cite{popat2018credeye} used stance detection to automatically predict the credibility of textual claims retrieved from the Web.
Another tool, developed by Nguyen et al. to assist fact-checkers~\cite{nguyen2018interpretable}, uses stance predicted from multiple articles alongside other attributes such as the reputation of the news sources to assess a claim's veracity.
Similarly, FAKTA~\cite{nadeem2019fakta}, a system for end-to-end fact-checking of claims, uses a stance detection model trained on FEVER setting alongside linguistic metadata for automatic fact-checking.

\descr{Remarks.} In this paper, we showed that \abbrv outperforms existing stance detection mechanisms and that it could be easily integrated into existing moderation and analysis pipelines that make use of stance detection.

\section{Discussion and Conclusion}
In this paper, we presented \approach, a new framework for detecting stance in social media posts discussing misleading claims.
We tested the ability of existing stance detection systems to aid content moderation on social media and identified three major limitations of these systems in the context of applying them for content moderation.
Motivated by the shortcomings, we developed the \abbrv framework as an unsupervised, platform, and topic-agnostic solution.
By experimenting on datasets from two social network platforms (Twitter and Reddit) and multiple topics (e.g., politics, health, climate) we showed that our method consistently outperforms both supervised and unsupervised baselines. 
Most importantly, we demonstrated that \abbrv can be easily integrated into an end-to-end content moderation system, improving the performance of the state-of-the-art soft moderation system Lambretta~\cite{paudel2022lambretta} by reducing its contextual false positives tenfold.

We believe that \abbrv will both serve as a new paradigm for unsupervised claim-based stance detection and will be a valuable tool for researchers and online platforms aiming to improve their existing content moderation systems.
Enabling context-aware soft moderation systems can go a long way in making our information ecosystems healthier, minimizing warning fatigue, and increasing the intended effectiveness of warning labels.
We now discuss the ethical concerns of our work, design implications and limitations of our approach, and avenues for future work.

\descr{Ethics.} 
All datasets used in this work were either publicly released by other researchers or were collected using publicly available APIs and following those API's terms of service.
This work is not considered human subjects research by our institution, since we do not interact with humans and do not collect any private information.
Nonetheless, we adhere to ethical standards by removing any personally identifying information when reporting examples of social media posts in the paper.
While we advocate that our approach should be used to benefit society, following the \emph{respect for public interest} and \emph{beneficence} principles of the Menlo report~\cite{kenneally2012menlo}, \abbrv could be misused by malicious parties.
Potential adversarial misuse includes biased platform providers using \abbrv to identify and downrank dissident users, or state-actors applying \abbrv to amplify false narratives of interest or identify expert accounts that are correcting/debunking false narratives to be silenced.
While these threats are real, automatically generating and posting content on social media at a large scale produces other artifacts that can be identified by alternative approaches~\cite{ikram2017measuring,nilizadeh2017poised,saeed2022trollmagnifier}.

\descr{Design implications.}
We envision \abbrv to be applied as a post-retrieval filtering tool for content moderation systems on social media after topically relevant candidate posts for a claim are retrieved, as illustrated in Figure~\ref{fig:lambretta_ctd}.
Since the approach of \approach is designed for the task of claim-specific stance detection, \abbrv is mostly intended for claim-specific content moderation systems like Lambretta~\cite{paudel2022lambretta}.
The process of integrating \abbrv as a tool to existing soft moderation systems is seamless, as we demonstrated in Section~\ref{sec:ctdwithlambretta}.
For each claim a platform moderator wishes to apply warning labels to, the only input from their end is to craft the triplet consisting of the \textcolor{blue}{consensus statement}, a \textcolor{red}{refuting evidence} and a \textcolor{ForestGreen}{supporting evidence}.
The triplet can then be used with the evaluation prompt in Figure~\ref{fig:prompt_setup} alongside the fine-tuned FLAN-T5 model for inference.
It is to note that one of the major advantages of using \abbrv is that it does not need any further fine-tuning for new claims and new platforms.
As demonstrated in the example triplets throughout our work (i.e. Figure~\ref{fig:triplets_eval} and Figure~\ref{fig:example}), the \textcolor{blue}{consensus statement} for a claim can be formulated with the most succinct fact-check or scientific consensus about the misleading claim.
Similarly, the contrastive markers can be simple statements that are positive affirmation and negative reframing of the same \textcolor{blue}{consensus statement}.
The final model was fine-tuned on normalized pieces of argumentative structures, and we expect it to be robust enough to handle different quality or phrasings of triplet semantic structure expected of a \abbrv triplet.

\descr{Limitations.}
There are some limitations that come with the task setup of \abbrv for stance detection.
First, we expect that any claim a platform aims to contextually moderate has been determined to be false, and has an accompanying fact-check statement associated with it.
This requirement of a corresponding fact-check statement used to build the \textcolor{blue}{consensus statement} in a \abbrv triplet poses a practical challenge in cases of quickly emerging false claims or novel skeptical narratives, for which a consensus about the claim has not yet been established.
For these cases, moderators could potentially use high-quality crowdsourced truth statements~\cite{allen2021scaling}, or resort to applying previously proposed soft moderation techniques that do not take stance into account (e.g., Lambretta).
The formulation of \abbrv can only handle posts that support or refute a misleading claim, while social media posts discussing misinformation and rumors might also contain posts that are of ``querying'' or ``commenting'' nature~\cite{zubiaga2016analysing,reyes2012humor}.
However, for the purpose of soft moderation where the objective is to apply warning labels to posts that are spreading the misleading claim, we argue that finer-grained distinction within the nature of ``support'' of a claim might not be necessary.
Upon manual inspection, we find that the majority of the misclassifications by \abbrv happen on posts that are sarcastic and satirical about the misleading claim being discussed, which are often misclassified as refuting the claim.
Identifying satire and sarcasm is a challenging NLP task~\cite{joshi2017automatic}, and future work can explore fine-tuning \abbrv models to handle more nuanced cases of stance occurring on social media text.

\descr{Future Work.}
In the future, we plan to extend \abbrv on evaluating claim-based stance detection in multi-lingual settings.
As LLMs become more powerful beyond the English Language, we can expect their learning capabilities to improve on multiple languages~\cite{scao2022bloom,wei2023polylm}.
We will also explore strategies to reduce the manual costs of curating the triplet structure needed for \abbrv by automatically matching the best set of fact checks and \textcolor{blue}{consensus statements} for misleading claims spreading in the wild, and generating the best set of triplets automatically.
The ClaimReview structured markup introduced by Google~\cite{claimreview} is a promising avenue for this direction, and we will explore multiple ways of semantically matching a misleading claim with ClaimReview markups, and generating the piece of contrastive markers from the fact-check document.

\descr{Acknowledgments.} We would like to thank the anonymous reviewers for their feedback.
This work was funded by the Institute for Global Sustainability and the Rafik B. Hariri Institute for Computing and Computational Science \& Engineering at Boston University under the \emph{Data and Misinformation in an Era of Sustainability and Climate Change Crises} project, and by the NSF under grants CNS-1942610, CNS-2114407, CNS-2247868, and DEB-2200052.

\bibliographystyle{plain}
\bibliography{refs.bib}
\end{document}